# Innovative 3D Depth Map Generation From A Holoscopic 3D Image Based on Graph Cut Technique

[1]Bodor Almatrouk, [2]Mohammad Rafiq Swash, [3]Abdul Hamid Sadka

Department of Electronic and Computer Engineering, College of Engineering, Design and Physical Sciences Brunel University London, Uxbridge, UB8 3PH, Great Britain of United Kingdom
me13baa@my.brunel.ac.uk[1], Rafiq.Swash@brunel.ac.uk[2], Abdul.Sadka@brunel.ac.uk[3]

*Abstract*—Holoscopic 3D imaging is a promising technique for capturing full-colour spatial 3D images using a single aperture holoscopic 3D camera. It mimics fly's eye technique with a microlens array, which views the scene at a slightly different angle to its adjacent lens that records three-dimensional information onto a two-dimensional surface. This paper proposes a method of depth map generation from a holoscopic 3D image based on graph cut technique. The principal objective of this study is to estimate the depth information presented in a holoscopic 3D image with high precision. As such, depth map extraction is measured from a single still holoscopic 3D image which consists of multiple viewpoint images. The viewpoints are extracted and utilised for disparity calculation via disparity space image technique and pixels displacement is measured with sub-pixel accuracy to overcome the issue of the narrow baseline between the viewpoint images for stereo matching. In addition, cost aggregation is used to correlate the matching costs within a particular neighbouring region using sum of absolute difference (SAD) combined with gradient-based metric and ―winner takes all‖ algorithm is employed to select the minimum elements in the array as optimal disparity value. Finally, the optimal depth map is obtained using graph cut technique. The proposed method extends the utilisation of holoscopic 3D imaging system and enables the expansion of the technology for various applications of autonomous robotics, medical, inspection, AR/VR, security and entertainment where 3D depth sensing and measurement are a concern.

*Keywords-holoscopic 3D image; integral image; elemental image; viewpoint image; depth map; disparity; disparity space image; sub-pixel accuracy; graph cut; holoscopic 3D camera*

## I. INTRODUCTION

Contemporary 3D imaging techniques provide significant benefits over conventional 2D imaging techniques and also most 3D technologies are competing to offer a new form of 3D objects that have natural colour, full parallax and without the need to wear any specific glasses [1]. Traditional 3D image system creates the illusion of depth by viewing two images simultaneously, all taken from slightly translated viewpoints of the same scene. This system is conventionally known as ―Stereoscopy‖ which mimics the human eye technique for both acquisition and visualisation. The stereoscopic technique uses two cameras that are slightly distanced from each other to capture images from different viewing angle, and this facilitates the perception of depth when the left eye image and right eye image are viewed by the left eye and right eye respectively However, images displayed by.]2[ such system are apt to causing eyestrain, fatigue, and headaches following prolonged viewing by the users [3]. Another limitation of using a stereoscopic system is the use of two cameras adds a high complexity to the system as well as depth perception because the cameras need to be accurately calibrated for different setup ,Furthermore.]33[ italso increases the cost and the size of the system and hence, it is not an ideal solution for dynamic applications.

With the emerging developments in digital imaging technology, demands for true 3D viewing yielded the development of more advanced 3D display ―autostereoscopic‖ systems [4]. These systems provide true 3D images and do not require 3D glasses, and consequently, are more comfortable to the user. Auto-stereoscopic technologies include holography [3], volumetric displays [5], multi-view [34] and holoscopic 3D imaging system also known as integral imaging [5,6]. All these systems combine stereo parallax and movement parallax effects to give 3D without glasses in multi-view and head-tracking auto-stereoscopic displays [3,7].

### A. Holoscopic 3D imaging principle

The concept of holoscopic 3D imaging system was first introduced in 1908 by Gabriel Lippmann [8] when he used natural light and a single-aperture camera with an array of micro-lenses to capture a scene to present fly's eye view artistically, which he called integral photography. Holoscopic technology provides motion parallax, binocular disparity, convergence [9] as well as an infinite refocus of the scene. Furthermore, holoscopic systems can create a true volume spatial optical model of the object scene in the form of planar intensity distribution [10].

In today's holoscopic 3D cameras, an array of microlenses is fitted behind the main lens and in front of the sensor to capture the colour and intensity of light rays from every direction, through every point in space, breaking down the entire image into individual rays of light. As a result, each micro-image also known as an elemental image, which the image captured behind a single micro-lens, preserves different information about the direction, colour, and intensity of light in the scene. Viewpoint images, which are images of the same scene from a different perspective, can be extracted from a single holoscopic image by reconstructing the pixels of the elemental images [11].

## II. RELATED WORK

Most research work conducted on Holoscopic 3D depth estimation techniques have been aimed at overcoming limitations related to H3DI reconstruction, and the determination of viewing parameters like depth of field

Kim et al. [12], and image quality of displayed images Martnez-Cuenca et al. [13]. The knowledge of holoscopic image depth or spatial position is one of the key focus in contemporary digital imaging applications [3,14,15], and its accuracy is used to improve a wide range of technical issues, such as coding and transmission. The advantages of 3D image depth determination in holoscopic digital image processing are closely related to the application of H3DI in fields like 3D cinema, robotic vision, medical imaging, detection and tracking of people, enhancing biometrics and in video games image generation [16,34]. The following are the current techniques that provide computational simulation of depth extraction in H3DI systems with micro-lens arrays:

Depth through disparity is an estimation of the relationship between image depth and disparity. Park et al. [15] and Wu et al. [6] were the first to document the testing of correlation metrics basing on disparity estimation. First, horizontal positions of the pixels present in the elemental images are rearranged to obtain the horizontal VPIs. To determine a method for extracting depth cues from H3DIs, the depth of an object point position is first derived through geometrical analysis of the H3DI recording using general depth estimation methods. A depth equation gives the relationship between depth and the corresponding displacement present between two 2D parallel records of a 3D space image associated with an H3DI recording [3].

The analysis of an object point Po (Xo, Z) is done by considering one lateral direction (x-direction), on a twodimensional lateral plane. The z-axis begins at the plane that is coincident with the micro-lenses' surface, whereas the x-axis is measured from the centre of the first microlens [1, 15]. The mathematical relationship representing depth and disparity is simplified as:

$$Z = (d.\varphi.f)/\Delta \quad (1)$$

where: $\Delta = ds_1 - ds_2$; is the sampling distance between two adjacent pixels, φ is the pitch size of the micro-lens sheet; and d is the disparity of the object point Po (Xo, Z) within two resampled pixels from two elemental images.

Depth through Anchoring Graph Cuts is another method for depth extraction. Zarpalas et al. [17] reports on extraction of reliable H3DI features, called ―anchor points‖ to cover for the limitations in the other methods. This depth estimation technique is an energy optimisation algorithm that improves depth estimation accuracy and reduces the complexity of the optimisation.

### III. 3D DEPTH MAP GENERATION

An efficient depth estimation algorithm based on calculating disparity map has been implemented to create a dense 3D model. Depth map extraction is measured from a single shot as a single holoscopic image consists of multiple viewpoint images. The viewpoints are extracted and utilised for disparity calculation via disparity space image (DSI) technique. Pixels displacement is measured with sub-pixel accuracy to overcome the issue of the narrow baseline between the viewpoint images for stereo matching. Cost aggregation is used to correlate the matching costs within a particular neighbouring region using Sum of absolute difference (SAD) combined with gradient-based metric and ―winner takes all‖ algorithm is employed to select the minimum elements in the array as optimal disparity value. Finally, the optimal depth map is obtained using Graph Cut algorithm.

### A. Viewpoint extraction

Given holoscopic raw image coordinate, the first plane coordinates *(u,v)* can be used to represent the viewpoint images. To extract the viewpoint images, the pixels from elemental images are rearranged to generate viewpoint images. From a particular *(u,v)* coordinate, the pixels sharing the same location *u* from each elemental image are extracted and arranged according to each pixel's associated micro-lens position *s* to generate a single viewpoint image. Thus, each elemental image is considered as one pixel of a viewpoint image. To compute the disparity, we have extracted the viewpoint images to utilise rather than the raw holoscopic images.

### B. Cost Computation

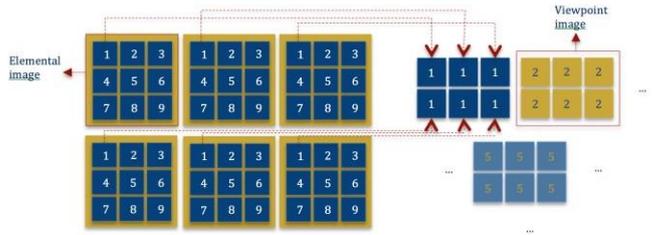

Figure 1. Extracting viewpoint images from a holoscopic 3D image by selecting pixels from elemental images

Given a calibrated and rectified viewpoint images, searching for corresponding points between a pair of viewpoint images is constrained by the epipolar line [18]. For each pixel lying on the epipolar line, disparity is computed, which is the distance between a pixel in one image and its corresponding pixel in the other image. The stereo matching algorithm is performed between the central viewpoint image as it has zero angle with the image plane [19] and the rest of the viewpoint images. To locate corresponding pixels, we have computed the matching costs using pixel-wise correlation over all pixels and all disparities, and stored those costs in a 3D volumetric structure defined as disparity space image (DSI) *D(x, y, d)*, where *D* is the cost function, *(x, y)* represents the pixel's coordinate and *d* denotes the disparity hypotheses [20]. As a result, DSI is constructed to evaluate the cost of different disparities.

There are various types of similarity measures to compute the matching costs; most notably squared differences (SD), absolute difference (AD), Normalised Cross-Correlation (NCC) as well as gradient-based measure [21]. In our algorithm, we have used the absolute difference, which is expressed by:

$$D_{AD}(u, d) = I(u_c, v) - I(u, v + d) \quad (2)$$

where $u_c$ denotes the central view image, $u = (s, t)$ represents the angular directions, $v = (x, y)$ represents the Cartesian image coordinates of each viewpoint image [22] and $d$ denotes the disparity hypotheses.

We have also employed gradient-based measure that is defined by:

$$D_G(u,d) = \delta(u) * \min(|\nabla_x I(u_c, v) - \nabla_x I(u, v + d)| \\ + \min((1 - \delta(u)) \\ * |\nabla_y I(u_c, v) - \nabla_y I(u, v + d)|) \quad (3)$$

where $\delta(u)$ denotes the effect of the directional differences regarding the relevant angular directions in *(s, t)* [23, 24].

Combining both metrics has previously shown a significant improvement in stereo matching algorithm [25]:

$$D(u,d) = D_{AD}(u,d)((1 - \alpha) \cdot \tau_1) + D_G(u,d)(\alpha \cdot \tau_2) \quad (4)$$

where $\alpha$ balances $D_{AD}(u, d)$ and $D_G(u, d)$ since the former applied on coloured image whilst the latter applied on grayscale image, and $\tau$ is a defined threshold that contains the outliers [26].

*C. Displacement measurement*

Applying the DSI algorithm directly on the viewpoint images will generate a poor result due to the narrow baseline between the viewpoint images. To overcome the issue of narrow baseline, we have computed pixel displacement with sub-pixel accuracy [27, 28] to attain sub-pixel precision while shifting the viewpoint images to compute the cost. Sub-Pixel accuracy displacement assures that the smallest translation is detected by applying phase correlation performed in the Fourier domain [29].

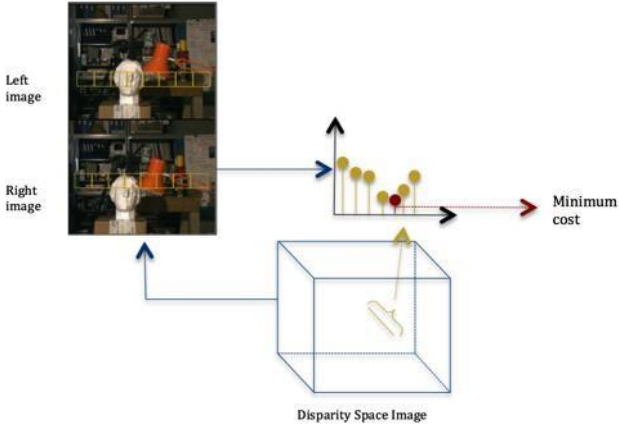

Figure 2. Disparity estimation process

Let two viewpoint images *I* shifted by $v_0$:

$$I(v_x) = I(v_x - v_{x0}) \quad (5)$$

The y direction is neglected as the matching is performed along the epipolar line (horizontally).

Since the displacement is performed in the Fourier domain, we calculate the 2D Fourier transform $\mathcal{F}$[26] of both original and shifted images:

$$\mathcal{F}\{I(v_x + \Delta v_x)\} = \mathcal{F}\{I(v_x)\}e^{2\pi i \Delta v_x} \quad (6)$$

The Phase Correlation function is achieved by applying the inverse of Fourier transform, which derived the Dirac function to be centred at $v_0$ to obtain a normalised crosscorrelation [29]:

$$I' \mathcal{F}^{-1}\{\mathcal{F}\{I(v_x)\}e^{2\pi i \Delta v_x}\} \quad (7)$$

By applying sub-pixel accuracy displacement equations to the similarity metrics, equations (2) and (3) are modified to:

$$D_{AD}(u,d) = \min(|I(u_c, v) - I(u, v + \Delta v_x(u, d)|) \quad (8)$$

$$D_G(u,d) = \delta(u) * \min(|\nabla_x I(u_c, v) \\ - \nabla_x I(u, v + d(u, v + \Delta v_x(u, \\ + (1 - \delta(u)) \\ * \min(|\nabla_y I(u_c, v) \\ - \nabla_y I((u, v + \Delta v_y(u, d))|) \quad (9)$$

where the cost is computed while shifting the viewpoint images at sub-pixel precision rather than the transformation of a local patch.

*D. Cost aggregation and disparity optimization*

Cost aggregation correlates the matching costs within a particular neighbouring region [29]. Pixel-wise cost calculation produces a noisy disparity map due to a low SNR (signal-to-noise ratio) [30]. As a result, cost aggregation compares costs over a fixed sized window instead of comparing single pixels, reducing the ambiguity (low SNR) that usually occurs when using pixel-by-pixel matching.

Computing the similarity using fixed sized window (kernel) is known as block-based stereo matching, where the window is shifted (in our case by sub-pixel) along the epipolar line and computing the similarity for each pixel pair using the aforementioned similarity metrics and choosing the minimum elements in array. This is known as winner takes all method. From this observation, we can reformulate equations (10) and (11) to:

$$D_{AD}(u,d) = \sum_{u,v} \min(|I(u_c,v) - I(u, v + \Delta v_x(u,d))|) \quad (10)$$

$$\begin{aligned}D_G(u,d) = &\sum_{u,v \in x} \delta(u) \\ &* \min(|\nabla_x I(u_c,v) \\ &\quad - \nabla_x I(u, v + d(u, v + \Delta v_x(u,d))|) \\ + &\sum_{u,v \in y}(1-\delta(u)) \\ &* \min(|\nabla_y I(u_c,v) \\ &\quad - \nabla_y I((u, v + \Delta v_y(u,d))|)\end{aligned} \quad (11)$$

A discrete set of disparities is to be chosen once the costs are computed. The disparities that have beenassigned to every pixel $p \in I$ are known as anchor points. Graph cuts estimate whether these anchor points are true scene points [31]. Assuming the set of disparity is $D$, the aim is to find a labelling $f$ that assigns each pixel a label $fp \in D$, where f is piecewise smooth and compatible with the observed data. The depth value of adjacent pixels should be taken into consideration, as they are significant to find the best solution for a pixel's depth value.

We have used OpenCV graph cuts based image segmentation algorithm [32] to achieve the optimal disparity map.

### IV. RESULT AND EVALUATION

The results of the proposed depth extraction algorithm were examined against DDFF 12-Scene dataset [35], which includes 720 holoscopic images and their depth maps. Each image consists of $9 \times 9 \times 383 \times 552$ viewpoint Images. Our results show cleaner and more defined edges whereas the dataset has shown coarser results see fig 3. The dataset contains noisy depth maps along with some incorrect disparity labelling as shown in fig 3, where our results provide a smoother, speckle-free and more detailed depth maps. However, the proposed algorithm has failed to accurately label the defocused areas in the scene.

To assess the effectiveness of the sub-pixel accuracy displacement, depth maps have been generated using both sub-pixel shift displacement and traditional pixel-by-pixel stereo matching. As shown in figure 3, the sub-pixel accuracy algorithm enhances the matching results tremendously compared to the pixel-by-pixel shift algorithm providing more details and accurate labelling.

Graph cut algorithm has also been evaluated. Depth maps generated without using graph cut are noisier and has large holes and speckles. The usage of graph cut eliminated large holes and speckles in the depth map as well as affecting the depth maps to be smoother while maintaining sharp edges and details.

### V. CONCLUSION

An innovative method for3D depth map extraction from a single holoscopic 3D image has been discussed. 3D depth maps were extracted from a single holoscopic 3D image by constructing a disparity space image that assigns a hypothesised disparity to each pixel derived from a subpixels matching in Fourier domain and selecting optimum disparity value through winner-takes-all cost aggregation method. The proposed method give robust and detailed depth maps that contain cleaner and more defined edges. The graph cut technique provided smoother and specklefree depth maps while sub-pixel displacement has eliminated the problem if narrow baseline and given accurate disparity labelling.

*A. Acknowledgement*

This publication was made possible by NPRP grant 9181-1-036 from the Qatar National Research Fund (a member of Qatar Foundation). The statements made herein are solely the responsibility of the authors.

REFERENCES

[1] A. E, A. A, A. Maysam, Swash. M.R, A. F. O, and F. J, "Scene depth extraction from Holoscopic imaging technology," IEEE, 2008, pp. 1–4.

[2] "How 3-D PC glasses work," HowStuffWorks, 2003.

[3] D. S. Pankaj, R. R. Nidamanuri, B. Pinnamaneni, and B. P. Prasad, "3-D imaging techniques and review of products," Sep. 2013.

[4] M.R. Swash, A. Aggoun, O. Abdulfatah, B. Li, J. C. Jacome, E. Alazawi, E. Tsekleves ―Pre-Processing of Holoscopic 3D Image For Autostereoscopic 3D Display‖, 5th International Conference on 3D Imaging (IC3D). 2013.

[5] A. Kubota, A. Smolic, M. Magnor, M. Tanimoto, T. Chen, and C. Zhang, "Multiview Imaging and 3DTV," *Signal Processing Magazine, IEEE*, vol. 24, no. 6, pp. 10 – 21, Nov. 2007.

[6] Aggoun, A.; Tsekleves, E.; Swash, M.R.; Zarpalas, D.; Dimou, A.; Daras, P.; Nunes, P.; Soares, L.D., "Immersive 3D Holoscopic Video System," MultiMedia, IEEE , vol.20, no.1, pp.28,37, Jan.March 2013.

[7] C. Connolly, "Stereoscopic imaging," *Sensor Review*, vol. 26, no. 4, pp. 266–271, Oct. 2006.

[8] "Épreuves Réversibles: Photographies Intégrales", *ComptesRendus de l'Academic des Sciences*, vol. 9, no. 146, pp. 446-451, 1908.

[9] M. Yamaguchi, "Full-Parallax Holographic Light-Field 3-D Displays and Interactive 3-D Touch", *Proceedings of the IEEE*, pp. 1-13, 2017.

[10] E. Alazawi, M. Swash and M. Abbod, "3D Depth Measurement for Holoscopic 3D Imaging System", *Journal of Computer and Communications*, vol. 04, no. 06, pp. 49-67, 2016.

[11] A. Vasquez Guzman, P. (2017). Generating Anaglyphs from Light Field Images. *Department of Mechanical Engineering Stanford University Stanford, CA*.

[12] J. Makanjuola, A. Aggoun , M. Swash , P. Grange , B. Challacombe, P. Dasgupta ―3D-Holoscopic Imaging: A Novel Way To enhance imaging in Minimally invasive therapy in urological oncology‖, Journal of Endourology. September 2012, 26(S1): P1A572.

[13] M.R. Swash, A. Aggoun, O. Abdulfatah, B. Li, J. C. Jacome, E. Alazawi, E. Tsekleves ―Pre-Processing of Holoscopic 3D Image For Autostereoscopic 3D Display‖, 5th International Conference on 3D Imaging (IC3D). 2013.

[14] J. C. Barreiro, M. Martínez-Corral, G. Saavedra, H. Navarro, and B. Javidi, "High-resolution far-field integral-imaging camera by double snapshot," *Optics Express*, vol. 20, no. 2, pp. 890–895, Jan. 2012.

[15] R. Ng, ―Digital light field photography,‖ 2006.

[16] E. Alazawi, M. Abbod, A. Aggoun, M. R. Swash, and O. Abdulfatah, ―Super Depth-map rendering by converting holoscopic viewpoint to perspective projection‖, 3DTV-CON in Pursuit of Next Generation 3D Display, Budapest, Hungary, 2-4th July 2014.

[17] E. Alazawi, A. Aggoun, O. Abdulfatah, M.R. Swash, ―Adaptive Depth Map Estimation from 3D Integral Images‖, IEEE International Symposium on Broadband Multimedia Systems and Broadcasting, London, UK, June 2013.

[18] N. Pears, Y. Liu and P. Bunting, *3D Imaging, Analysis and Applications*, 1st ed. London: Springer London, 2012, pp. 35-94

[19] D. Zarpalas, E. Fotiadou, I. Biperis and P. Daras, "Anchoring Graph Cuts Towards Accurate Depth Estimation in Integral Images", *Journal of Display Technology*, vol. 8, no. 7, pp. 405-417, 2012.

[24] W. Shao, H. Sheng and C. Li, "Segment-Based Depth Estimation in Light Field Using Graph Cut", *Knowledge Science, Engineering and Management*, pp. 248-259, 2015.

[25] A. Klaus, M. Sormann and K. Karner, "Segment-Based Stereo Matching Using Belief Propagation and a Self-Adapting Dissimilarity Measure", *18th International Conference on Pattern Recognition (ICPR'06)*, 2006.

[26] P. Tan and P. Monasse, "Stereo Disparity through Cost Aggregation with Guided Filter", *Image Processing On Line*, vol. 4, 2014.

[27] C. Kuglin and D. Hines, "The phase correlation image alignment method", *Proc. Int. Conference on Cybernetics and Society*, pp. 163–165, 1975.

[28] E. Vera and S. Torres, "SUBPIXEL ACCURACY ANALYSIS OF PHASE CORRELATION REGISTRATION METHODS APPLIED TO ALIASED IMAGERY", *16th European Signal Processing Conference (EUSIPCO 2008), Lausanne, Switzerland*, 2008.

[29] H. Hirschmuller, "Stereo Processing by Semiglobal Matching and Mutual Information", *IEEE Transactions on Pattern Analysis and Machine Intelligence*, vol. 30, no. 2, pp. 328-341, 2008.

[20] M. Kim, T. Oh and I. Kweon, "Cost-aware depth map estimation for Lytro camera", *2014 IEEE International Conference on Image Processing (ICIP)*, 2014.

[21] S. Unnikrishnan, S. Surve and D. Bhoir, *Advances in Computing, Communication and Control*, 1st ed. Berlin, Heidelberg: Springer Berlin Heidelberg, 2011, pp. 522-529.

[22] T. Georgiev and A. Lumsdaine, "Reducing Plenoptic Camera Artifacts", *Computer Graphics Forum*, vol. 29, no. 6, 2010.

[23] Jingyi Yu, L. McMillan and S. Gortler, "Scam light field rendering", 10th Pacific Conference on Computer Graphics and Applications, 2002. Proceedings..

[30] M. Habib and J. Davim, Interdisciplinary Mechatronics: Engineering Science and Research Development, 1st ed. John Wiley & Sons, 2013, pp. 483-498.

[31] "Miscellaneous Image Transformations — OpenCV 2.4.13.2 documentation", *Docs.opencv.org*, 2017. [Online]. Available: http://docs.opencv.org/2.4/modules/imgproc/doc/miscellaneous_transformations.html#grabcut. [Accessed: 02- May- 2017].

[32] "OpenCV library", *Opencv.org*, 2017. [Online]. Available: http://opencv.org/. [Accessed: 05- Sep- 2017].

[33] A. Wilson, "Choosing a 3D vision system for automated robotics applications," in *Vision Systems*, 2014.

[34] Y. Zhang, Q. Ji, and W. Zhang, ―Multi-view autostereoscopic 3D display,‖ in Int. Conf. on Opt. Photon. Energy Eng. (OPEE), 2010.

[35] Hazırbaş, Caner & Leal-Taixé, Laura & Cremers, Daniel. (2017). ―Deep Depth From Focus‖, Arxiv preprint arxiv:1704.01085.

| Holoscopic 3D Images | 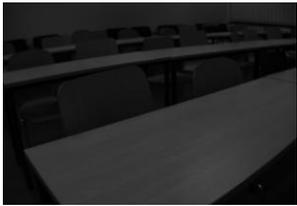 | 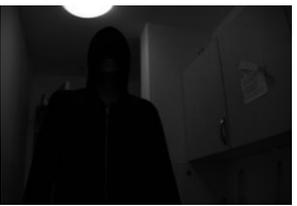 | 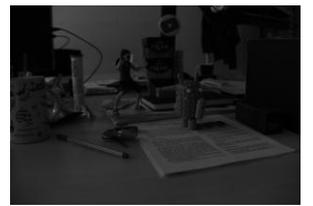 |
|---|---|---|---|
| A patch of zoom view holoscopic 3D images | 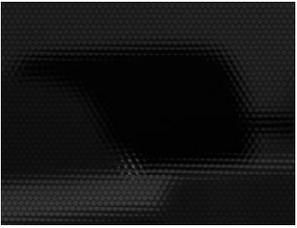 | 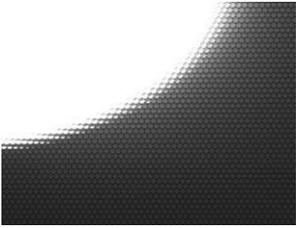 | 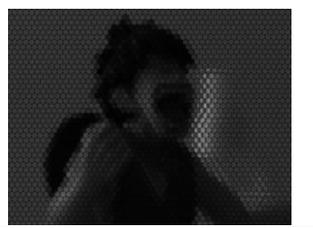 |
| Viewpoint images | 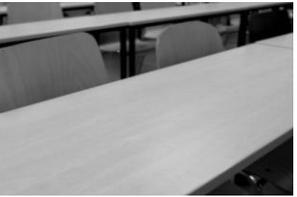 | 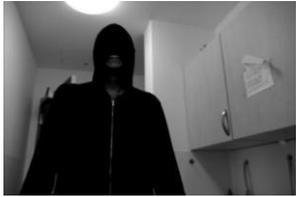 | 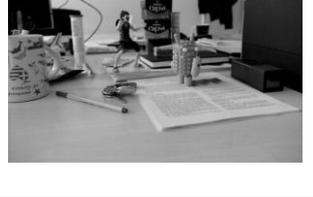 |
| State of the art method deep depth from focus (DDFF) [35] | 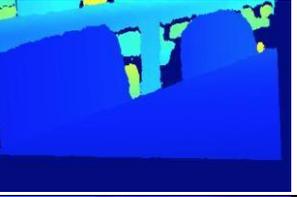 | 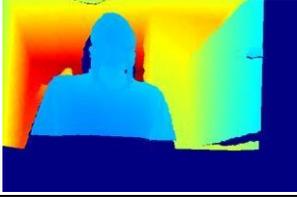 | 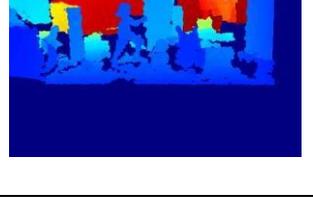 |
| The proposed method Without graph cut technique | 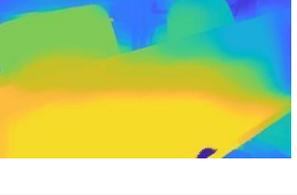 | 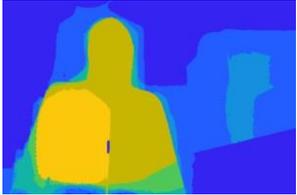 | 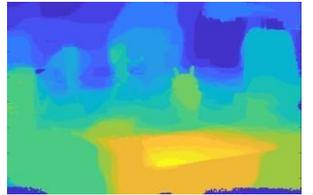 |
| The proposed method with pixel-by-pixel shift | 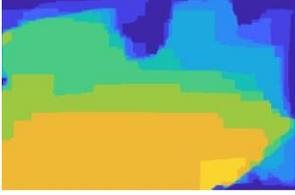 | 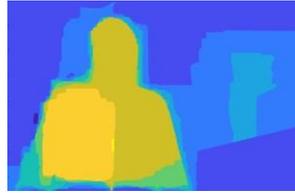 | 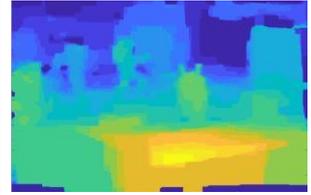 |
| The Proposed Method with graph cut technique | 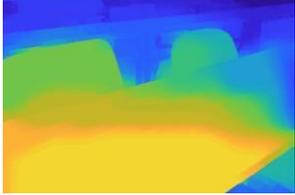 | 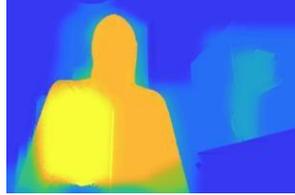 | 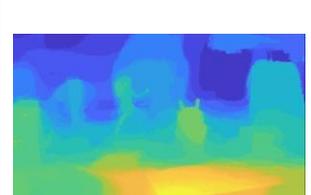 |

Fig. 3. 3D Depth map generation and performance with like-to-like comparision